\AtBeginDocument{%
  \providecommand\BibTeX{{%
    \normalfont B\kern-0.5em{\scshape i\kern-0.25em b}\kern-0.8em\TeX}}}

\documentclass[sigconf]{acmart}
\usepackage{cite}
\usepackage{amsmath,amsfonts}
\usepackage{algorithmic}
\usepackage{graphicx}
\usepackage{textcomp}
\usepackage{appendix}
\usepackage{tikz}

\copyrightyear{2022}
\acmYear{2022}
\setcopyright{acmcopyright}\acmConference[ISWC '22]{The 2022 International Symposium on Wearable Computers}{September 11--15, 2022}{Cambridge, United Kingdom}
\acmBooktitle{The 2022 International Symposium on Wearable Computers (ISWC '22), September 11--15, 2022, Cambridge, United Kingdom}
\acmPrice{15.00}
\acmDOI{10.1145/3544794.3558463}
\acmISBN{978-1-4503-9424-6/22/09}

\acmConference[ISWC '22]{International Symposium on Wearable Computers}{September 11 — 15
}{Atlanta, USA}
  
\begin{document}
\title{Shape Analysis for Pediatric Upper Body Motor Function Assessment}


\author{Shashwat Kumar}
\affiliation{
\institution{Systems Engineering, UVA}
\city{Charlottesville}
\country{USA}
}
\email{sk9epp@virginia.edu}

\author{Robert Gutierez}
\affiliation{
\institution{Systems Engineering, UVA}
\city{Charlottesville}
\country{USA}
}
\email{rjg7ra@virginia.edu }

\author{Debajyoti Datta}
\affiliation{
\institution{Systems Engineering, UVA}
\city{Charlottesville}
\country{USA}
}
\email{dd3ar@virginia.edu}

\author{Sarah Tolman}
\affiliation{
\institution{Systems Engineering, UVA}
\city{Charlottesville}
\country{USA}
}
\email{sat2ew@virginia.edu}

\author{Allison McCrady}
\affiliation{
\institution{Biomedical Engineering, UVA}
\city{Charlottesville}
\country{USA}
}
\email{anm9xd@virginia.edu}

\author{Silvia Blemker}
\affiliation{
\institution{Biomedical Engineering, UVA}
\city{Charlottesville}
\country{USA}
}
\email{ssblemker@virginia.edu}

\author{Rebecca J. Scharf, MD, MPH}
\affiliation{
\institution{Neurodevelopmental Pediatrics, UVA}
\city{Charlottesville}
\country{USA}
}
\email{rs3yk@hscmail.mcc.virginia.edu}

\author{Mahdi Boukhechba}
\affiliation{
\institution{Systems Engineering, UVA}
\city{Charlottesville}
\country{USA}
}
\email{mob3f@virginia.edu}

\author{Laura Barnes}
\affiliation{
\institution{Systems Engineering, UVA}
\city{Charlottesville}
\country{USA}
}
\email{lb3dp@virginia.edu}

\renewcommand{\shortauthors}{Kumar et al.}

\begin{abstract}
Neuromuscular disorders, such as (SMA) and Duchenne Muscular Dystrophy (DMD), cause progressive muscular degeneration and loss of motor function for 1 in 6,000 children. Traditional upper limb motor function assessments do not quantitatively measure patient-performed motions, which makes it difficult to track progress for incremental changes. Assessing motor function in children with neuromuscular disorders is particularly challenging because they can be nervous or excited during experiments, or simply be too young to follow precise instructions. These challenges translate to confounding factors such as performing different parts of the arm curl slower or faster (phase variability) which affects the assessed motion quality. This paper uses curve registration and shape analysis to temporally align trajectories while simultaneously extracting a mean reference shape. Distances from this mean shape are used to assess the quality of motion. The proposed metric is invariant to confounding factors, such as phase variability, while suggesting several clinically relevant insights. First, there are statistically significant differences between functional scores for the control and patient populations (p$=$0.0213$\le$0.05). Next, several patients in the patient cohort are able to perform motion on par with the healthy cohort and vice versa. Our metric, which is computed based on wearables, is related to the Brooke's score ((p$=$0.00063$\le$0.05)), as well as motor function assessments based on dynamometry ((p$=$0.0006$\le$0.05)). These results show promise towards ubiquitous motion quality assessment in daily life.
\end{abstract}
\maketitle

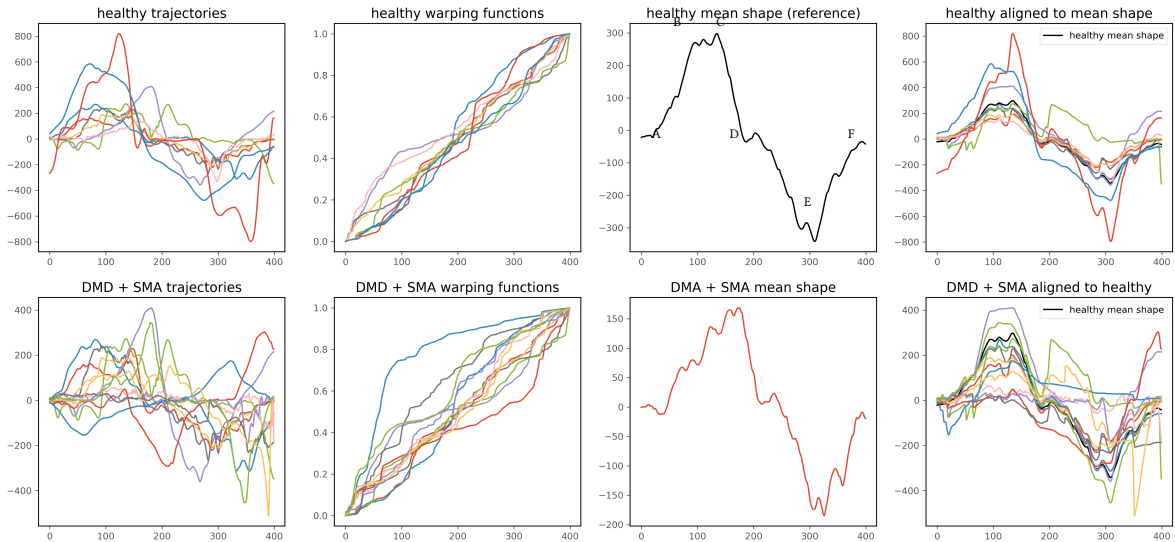
\begin{figure*}
\begin{tikzpicture}
\pgftext{%
\includegraphics[width=0.88\textwidth]{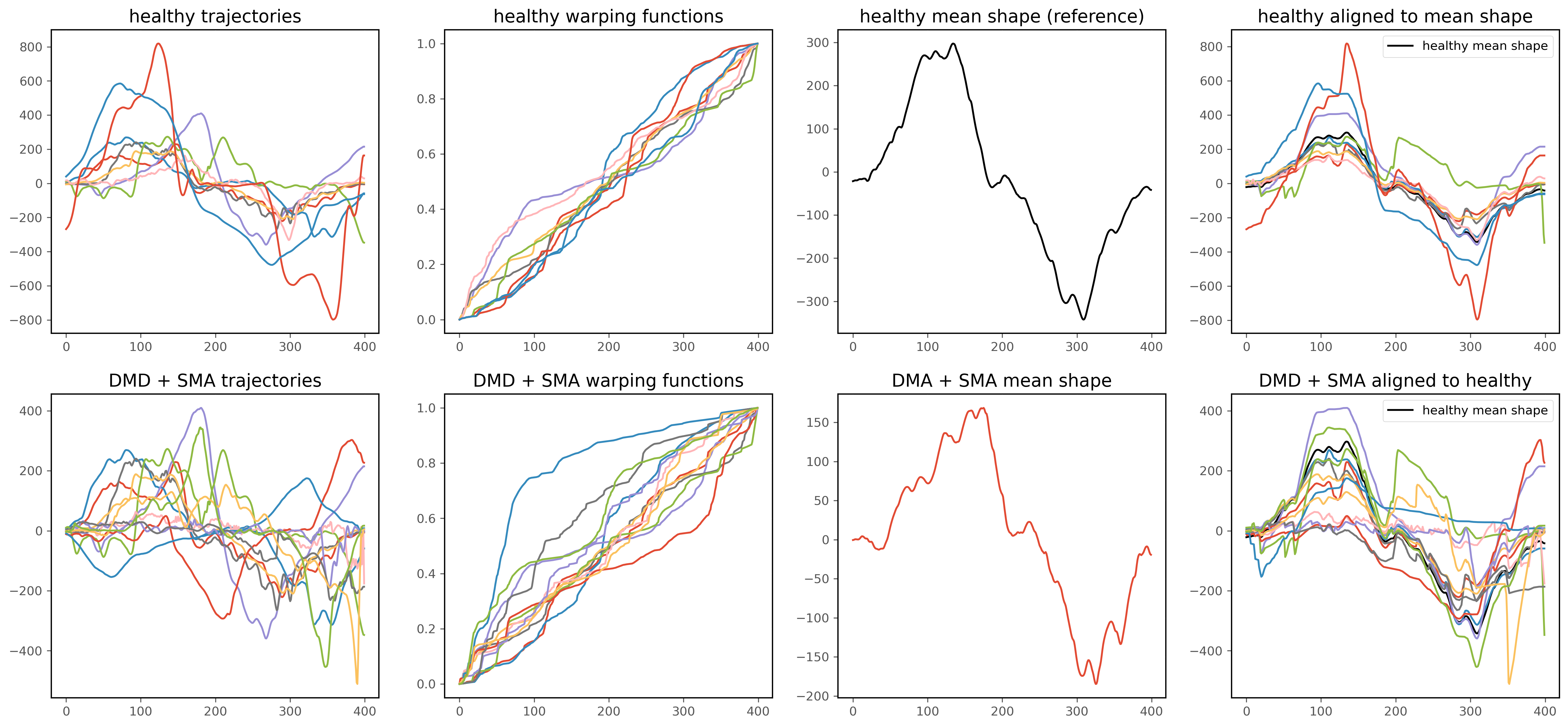}%
\node at (-7.3,5.5) {\tiny A};
\node at (-7.1,7) {\tiny B};
\node at (-6.6,7) {\tiny C};
\node at (-6.5, 5.5) {\tiny D};
\node at (-5.6,4.6) {\tiny E};
\node at (-5.1,5.5) {\tiny F};
}
\end{tikzpicture}
\caption{Top row: Given a set of healthy trajectories, we find a set of temporal warping functions which align the trajectories with each other, allowing us to find the elastic mean for healthy cohort (black). Bottom row: We align DMD + SMA trajectories to the elastic mean of healthy cohort. The warping functions corresponding to DMD + SMA are more distorted than the ones corresponding to healthy cohorts. Based on the healthy elastic mean, we compute three distances: amplitude (based on difference in y values), phase: based on differences in warping functions, cosine: based on cosine similarity between the aligned functions.}
\label{fig:phase_amplitude}
\end{figure*}

\section{Introduction}
While there has been significant progress in the development of therapeutic treatments for two prominent neuromuscular disorders, Spinal Muscular Atrophy (SMA) and Duchenne Muscular Dystrophy (DMD), the assessments for measuring patient progress have remained stagnant \citep{ottesen2017iss}. SMA affects approximately 1 in 11,000 babies born each year, with a life expectancy at just under 2 years while DMD affects 1 in 3,500 males. These new therapeutics have increased the life expectancy of patients with SMA from infants to young adults, and from early teens to mid thirties for patients with DMD. For boys with DMD, they typically lose their ability to walk in late childhood to early teen years, and have limited use of their arms after their early teens.  

Currently, there are several motor function assessments that physicians use to measure a patient's functional abilities within a clinical setting, such as the Children's Hospital of Philadelphia Infant Test of Neuromuscular Disorders (CHOP-INTEND) \citep{Glanzman2010TheReliability}, the Brooke Upper Extremity Scale \citep{florence1984clinical}, and the Hammersmith Functional Motor Scale Expanded (HFMSE) \citep{Ramsey2017RevisedTool}. Unfortunately, all the assessments previously mentioned use outcomes based on ordinal, subjective rating scales, which are unable to account for incremental changes, for both quantity and quality of movement. Another problem with data collected through clinical visits is related to sparsity. Due to the nature of the neuromuscular disorder, patient visits are infrequent, usually every 4-6 months. Thus, the low sampling frequency makes it challenging to gauge the temporal dynamics of the neuromuscular disorder. This is another avenue in which motion quality assessment approaches based on wearables can prove to be useful, since the assessment can in principle be done in a home setting by integrating ubiquitous sensors in activities of daily living (ADLs).


However, pediatric motion quality assessment is challenging because of several confounding factors, depending upon the modality used. Firstly, translations of the device (eg. moving the kinect/camera closer or further away from the subject) can affect the trajectory representations based on spatial positions. Similarly, rotations of either the participant or the sensor can change representations based on (x,y,z) coordinates \citep{amor2015action}. Variations in limb length affect all representations based on 3d-position.  
Finally, phase variation (for instance, choosing to do the first half of the curl slower than the last) tends to affect data collected from wearable devices \citep{Choi2018TemporalData}. 

An example of this phase variability can be seen in Figure \ref{fig:phase_amplitude}, first column. Here, several examples of arm curls are shown for both healthy and patient cohort. There is significant phase variability in the healthy cohort with peaks and valleys for different participants occurring at different times. This variability makes it challenging to compare and analyze these trajectories. Furthermore, to temporally align these trajectories, we need to first establish a reference trajectory against which to align them. One good candidate is the mean of these trajectories; however we first need to find a mean shape for the trajectories, which first requires temporal alignment to find the reference.

In this work, we demonstrate that this problem can be cast naturally in the language of differential geometry and shape analysis \citep{klassen2004analysis, fletcher2007riemannian, kendall2009shape,amor2015action}. First, by working with joint angles in our representation, we avoid variations due to limb length, which is especially important considering our participants have a wide variation in age and size. Then, we perform phase amplitude separation, an iterative procedure to simultaneously discover a mean shape of the trajectories while temporally aligning the trajectories with this mean shape. Next, we construct three distances from this mean shape based on amplitude, phase differences and cosine dissimilarity. In our work, we illustrate that by using the distances between the aligned curves, clusters in the data become apparent which leads to more sensible classification. We also demonstrate that amplitude distances are related to motor function assessed from Brooke's score as well as from dynamometry. To the best of our knowledge, this is the first work using shape analysis for neuromuscular motion quality assessment. We now provide a brief review of the our methods, our results as well as their future implications.

\section{Experimental Protocol}
Through the Pediatric Neuromuscular Clinic at the University of Virginia Children's Hospital, data was collected from 41 participants. Patients were either diagnosed with SMA or DMD and control participants were recruited based on the enrolled patients, matching both age and sex ($\pm$ 1 year). We placed a MetaMotionR+ (MbientLab, San Francisco, CA, USA) sensor on the top of each hand, with the accelerometer and gyroscope sensors sampled at 200 Hz. Then we had each subject perform a series motions related to activities of daily living, such as turning a door knob, raising a cup to their mouth, etc. The Brooke Upper Extremity Scale was used to provide one standardized metric for comparison between all cohorts \citep{Brooke1981ClinicalProtocol}. The study was approved by the University of Virginia's Internal Review Board for Health Sciences Research, protocol \#12161. For this study, some patients' data has been excluded from the subsequent analysis for the following reasons: 1. sensor malfunction (2); 2. refusal to cooperate due to young age (2); 3. patient withdrew from study (1). 

\begin{figure*}
    \centering
    \includegraphics[width=1.0\textwidth]{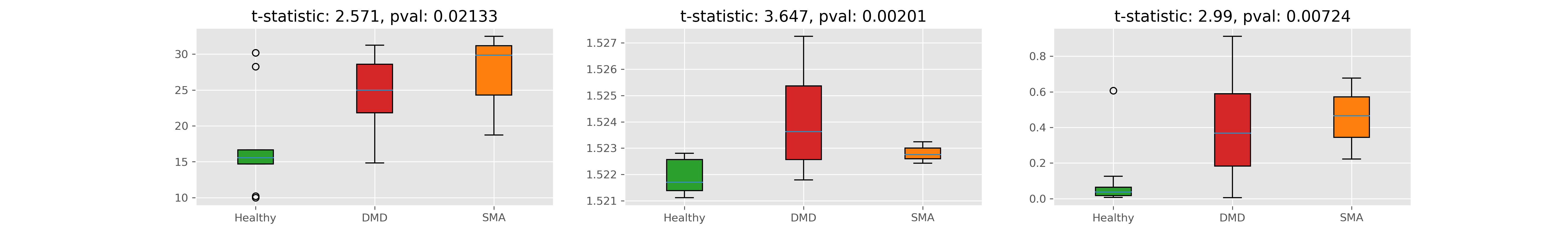}
    \caption{Boxplots corresponding to amplitude, phase and cosine distances. We run a t-test with unequal variances, comparing healthy cohort with DMD + SMA. In all three distances, DMD+SMA distances distances are statistically different from the healthy ones.}
    \label{fig:shape_distribution_plots}
\end{figure*}

\begin{figure*}
    \centering
    \includegraphics[width=0.45\textwidth]{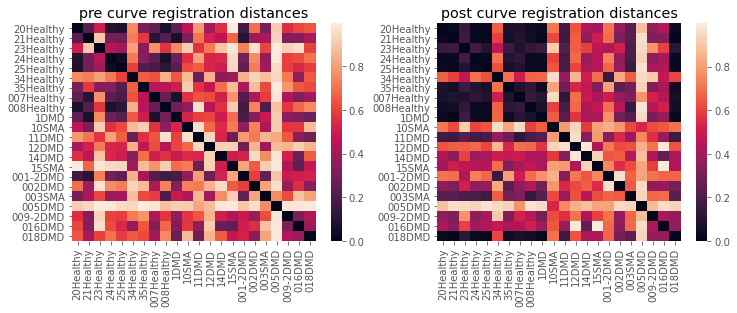}
    \includegraphics[width=0.25\textwidth]{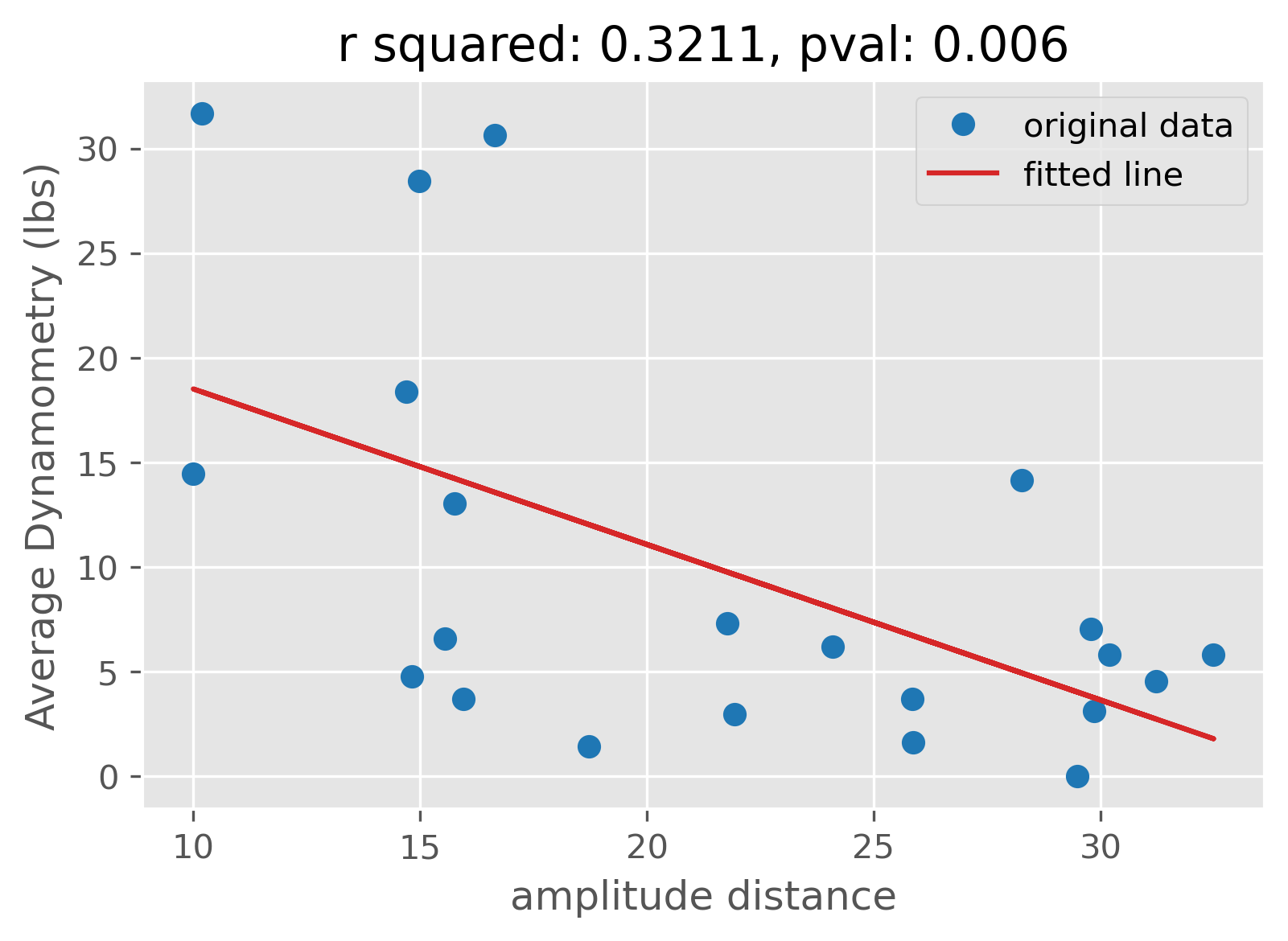}
    \includegraphics[width=0.25\textwidth]{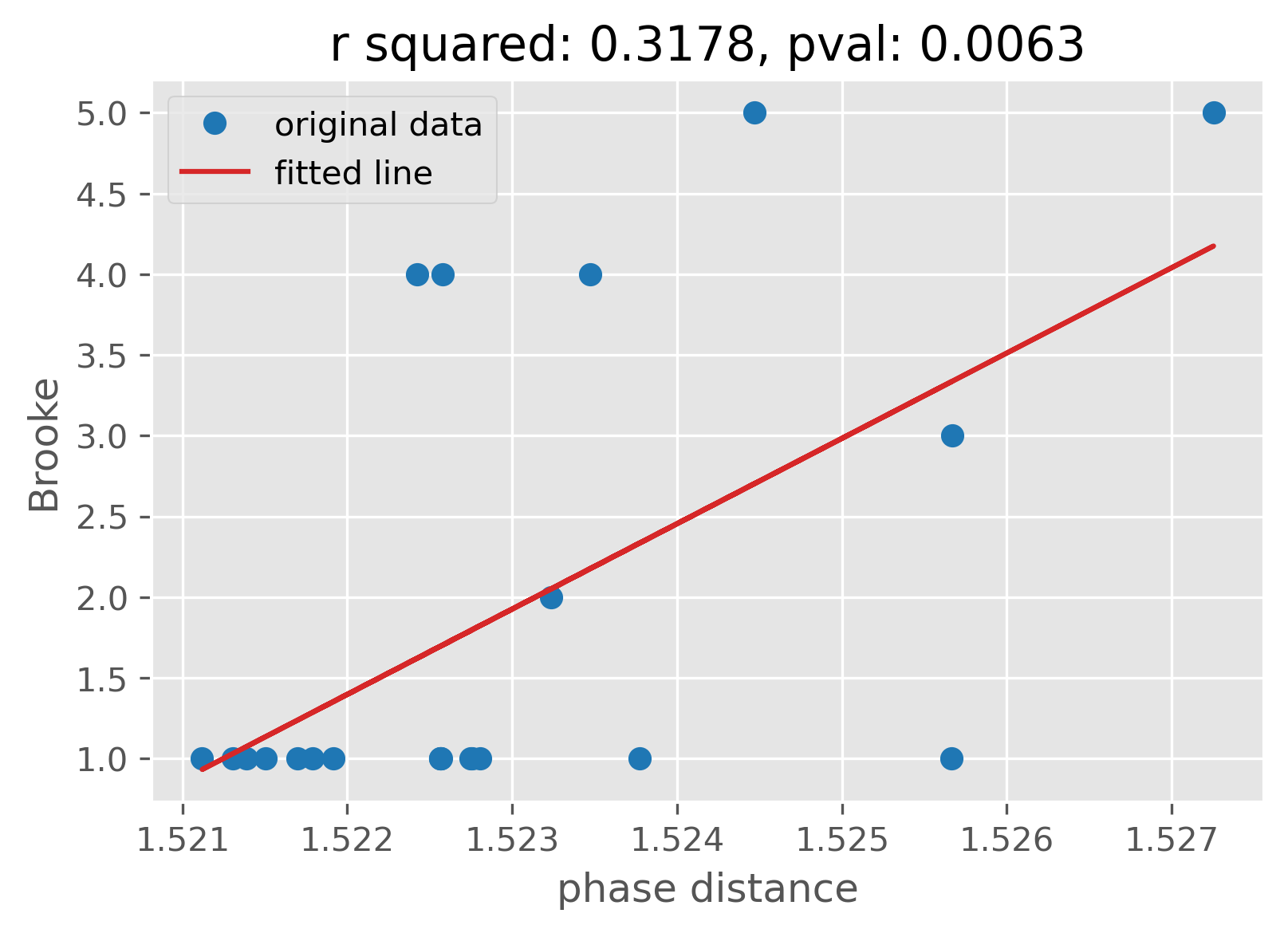}
    \caption{a,b) Distance matrices pre and post performing curve registration. The registered curves seem to display a much nicer block structure. Several patients of the DMD + SMA cohort perform motion on par with healthy. Participant 34 performs motion much more worse than other healthy patients. c) Amplitude distance is linearly related with muscle score 1 captured via  handheld dynamometer. The phase distance is related to Brooke's score, the current gold standard for functional assessment.}
    \label{fig:distance_matrices}
\end{figure*}

\section{Methodology}
Let $\beta_{t}$ be the function we are interested in modeling. In our case, it represents the gyroscope's signal. Given raw data, we resample, smooth, and find its derivative to first obtain $\dot{\beta}(t)$. Our initial preprocessing step for reducing noise in the data included the use of a 3rd order, low-pass Butterworth filter with a cutoff frequency of 0.1 Hz.

The srvf (\citep{klassen2004analysis}) of a function is then defined as:
$ q_t = \frac{\dot{\beta_t}}{\sqrt{\lVert \dot{\beta_t} \rVert}}  $.

This allows us to define the following two distances:

$
    A(\beta_1(t), \beta_2(t)) = inf_{\gamma} \lVert q_1 - q_2 \circ \gamma \sqrt{\dot{\gamma}} \rVert 
$ 

Similarly, after finding the optimal warping function which aligns two functions, the phase distance between them is given by:

$
   P(\gamma_{a\rightarrow b}) = cos^{-1}(\langle \gamma_{id}, \sqrt{\dot{\gamma_{a \rightarrow b}}} \rangle) 
$

In order to compute the pairwise distance matrices because some of the trajectories were particularly noisy, we also compute the cosine distance between the aligned functions 

$
    C(\beta_1(t), \beta_2(t)) = 1.0 - cos^{-1} (\langle \beta_1(t), \beta_t(t) \rangle)
$

Once the srvf has been defined, we can perform phase amplitude separation as outlined in \citep{tucker2013generative} to align the trajectories and also calculate the mean shape. We use the python package $fdasrsf$ for our analysis.


\section{Results}

In Figure \ref{fig:phase_amplitude}, we show the results of performing phase amplitude separation on our dataset. The first group comprises a set of healthy trajectories, which is shown in the top left plot. The second group comprises of DMD + SMA, whose trajectories are shown below. As we can see from the raw trajectories and the healthy cohort in particular, the data has a lot of phase variability meaning that despite having the similar shapes, the peaks and valleys occur at different times in different trajectories.  We now run phase-amplitude separation on the set of healthy trajectories, as described in the section above. This aligns these functions temporally while simultaneously discovering a mean shape. The elastic mean shape of healthy arm curls is shown in the third plot, while the corresponding temporal warping functions are shown in the second plot. 

The mean shape has a rather interesting structure. It consists of a stationary phase when the curl is initiated (A). The peak angular velocity is not reached immediately but through transient stage (B) which involves a slight deceleration after which peak velocity is reached (C). From looking at the videos, we discovered that the twin peaks (B to C) is because of a slight flicking of the wrist toward the shoulder.  What's even more interesting is that the mean shape does not smoothly decelerate to 0, first overshooting it a little (D) and then coming back to 0. Similarly, in the deceleration phase, involves two peaks  (E) followed by return to  0 (F). The symmetrical shape of the motion indicates that the participant performed the same amount of rotation up and down. The warping functions provide a sense of how much work needs to be done in order to map each function to the healthy mean. From the top right plot, we can see that the peaks and valleys of the healthy trajectories align quite nicely with the mean shape.

\begin{figure*}
    \centering
    \includegraphics[width=0.99\textwidth]{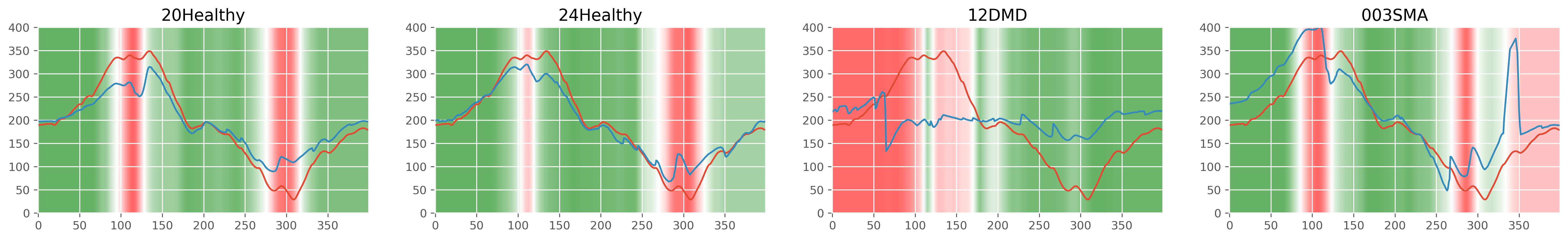}
    \caption{Assessing motion quality over different phases of the arm curls.}
    \label{fig:rolling_corr}
\end{figure*}

In the second row, the corresponding DMD + SMA trajectories are aligned temporally to the elastic healthy mean. The first plot shows the raw data while the second plot shows the warping needed in order to align the unhealthy trajectories with healthy. In the rightmost plot, we overlay the aligned trajectories on the elastic mean. Here we see much poorer agreement between the peaks and valleys of the DMD + SMA cohort and the elastic mean. As we can see visually, the warping functions for the DMD + SMA cohorts are a lot more distorted, implying that the DMD + SMA cohorts need to be stretched/squished a lot more in order to align them to the healthy mean compared to the warping needed for healthy trajectories. We also calculate the elastic mean corresponding to the DMD + SMA cohort. Because of the large variation in trajectories, this elastic mean is much more noisier.  

We compute the elastic phase and amplitude distance to the healthy mean. The distribution of the aforementioned distances is shown in Figure \ref{fig:shape_distribution_plots}. As we can see, the DMD + SMA cohorts have larger amplitude, phase and cosine distances compared to healthy mean. We run a t-test with unequal variance assumption to compare DMD + SMA with healthy. For amplitude distances, the test statistic is 2.571 and the p-value is 0.02133. For phase distances, the test statistic is 3.647, while the p value is 0.00201 Finally, for the cosine distances, the test statistic is 2.99 with a p value of 0.00724. Thus both amplitude, phase and cosine distances from the registered mean seem to capture differences between the healthy and non healthy cohorts. In all three cases, the healthy cohort had much lesser variance than DMD. Another subtlety we noticed was that the amplitude distance had an extra outlier (participant 17) which was not present in the phase and cosine similarity distances. We suspect this is because participant 17 had performed the motion almost 3 times as fast as the other ones. The amplitude distance being based in frobenius norm is more sensitive to larger magnitudes while the cosine dissimilarity (and also the phase distance) because of its normalization term is less susceptible to such larger velocities.


In order to establish the utility of curve registration, In Figure \ref{fig:distance_matrices} a,b, we plot pairwise cosine distances between points pre and post curve registration. We used cosine distance instead of the elastic distance as some of the DMD+SMA trajectories are really noisy and elastic distance involves an alignment step which performs poorly with the noisy trajectories. The participants have been ordered by cohort with all the healthy ones being close to each other. As we can see, the distance matrix on the left does not have a visually discernible block structure of any sort, which would indicate clustering. This is not surprising since the presence of phase variability in data obfuscates some of the structure present. The distance matrix post curve registration seems to have a clear block structure with the set of healthy controls having much smaller distances from each other. Interestingly, some of the participants with DMD (18 and 1) seem to have smaller distances than the healthy participants despite belonging to the DMD class. It's also surprising that one of the healthy participants (34) seems to have much larger distances than the non healthy cohorts.

The aim of this work  is to develop an index of symptom severity based on unsupervised motion data. To do so, we compare the relationship of the indices: amplitude distance, and phase distance with the Brookes score as well as the patient strength measured from dynamometry  (Figure \ref{fig:distance_matrices} c).
The amplitude distance had a statistically significant linear relationship with the dynamometry index (\textbf{p$=0.006 \leq 0.05$}). This makes sense because patients with larger muscle strength were able to better perform the motion and have a lower amplitude distance. The phase distance has a statistically significant linear relationships with the Brooke's score which is the current clinical gold standard for assessing motor function ((\textbf{p=$0.0337 \leq 0.05$})). Larger phase distances imply larger Brookes score, indicating greater symptom severity.

These preliminary results show promise towards using wearable technologies to capture movement patterns and daily life and augment time-consuming, less frequent physiological assessments requiring expensive lab grade equipment and subjective measures like the CHOP-INTEND.
The long-term goal of this work is to develop an index for assessing motion quality over time. While we do not currently have the longitudinal data to do so, we identify salient patterns across our healthy controls, DMD, and SMA participants. In Figure \ref{fig:rolling_corr}, we plot the rolling correlation between the reference aligned signal with the healthy mean. Participants 20 and 24  from the healthy cohort perform the arm curl motion very similar to the elastic mean, with differences at the peak and valley. The difference at the peak translates to how much of a wrist flick occurred when reaching the shoulder, while the valley is indicative of whether they returned their arm to the original starting position, or continued downward to their leg. Participant 12 with DMD on the other hand, seems to not be able to perform the first part of the motion at all, which was expected due to their limited muscle strength. Participant 3 with SMA does a good job in the first part of the curl, while struggling with the last part of the curl, as they had difficulty rotating back to the original starting position.

    


\section{Discussion}
In this work, we presented evidence for how wearables coupled with shape analysis can allow us to come up with a metric for assessing motion quality. Phase amplitude separation allowed us to overcome issues with phase variability so we can capture functional differences between the healthy and patient cohorts. Moreover, our metric was able to detect subtle differences for some of the non healthy participants that would have otherwise gone unnoticed. 
The rolling correlation between registered trajectories allowed us to understand when parts of the motion were performed well and when it deviated from the accepted path. We also discovered a statistically significant relationship between our functional muscle scores and the muscle score obtained from ultrasound and dynamometry.

Our results have several important implications: One can imagine using such a system for home use, where we can track data from patients with neuromuscular disorders over longer durations and without burdening the patient to come into the clinic multiple times. This will not only allow doctors to collect more data, but also aid doctors in tracking progress that is often noted from patient caregivers. By coupling such a system with activity recognition classifiers, it's possible for a clinician to monitor function progress in a variety everyday activities.  Finally, the non intrusive nature of the monitoring system makes it possible to track longitudinal progression, which is especially important for patients who have received revolutionary treatments like gene therapy. Telemedicine with wearable sensors will allow clinicians to offer support to patients during events like pandemics and more remote areas of the world.

We recognize that this is a very limited sample size compared to similar work; however, recruitment for this population is fairly limited, and thus, our cohort size is rather impressive for this patient population. We understand  that the quality of motion need not be a single dimension as patients may be compensating for muscular disorders by performing motions in a particular manner, something we are investigating further. 

\textbf{Funding}: This work was funded in part by a University of Virginia Engineering in Medicine Award.



\bibliographystyle{ACM-Reference-Format}
\bibliography{references}

\end{document}